\pdfoutput=1

\documentclass[11pt]{article}
\usepackage{ACL2023}
\usepackage{times}
\usepackage{latexsym}
\usepackage[T1]{fontenc}
\usepackage[utf8]{inputenc}
\usepackage{microtype}
\usepackage{inconsolata}
\usepackage{svg}
\usepackage{amsfonts}
\usepackage{amssymb}
\usepackage{subcaption}
\usepackage[english]{babel}
\usepackage{amsmath}
\usepackage{float}
\usepackage{graphicx}
\usepackage{amsmath,amssymb,pgf}
\usepackage{hyperref}
\usepackage{color, colortbl}

\newcommand*{\parallelogramm}{%
  \rlap{\rotatebox{-30}{\rule[.05ex]{.4pt}{.77em}}}%
  \kern.04em%
  \rlap{\kern.36em\raisebox{0.649519052835em}{\rule{.6em}{.4pt}}}%
  \rule{.6em}{.4pt}\kern-.04em%
  \rotatebox{-30}{\rule[.05ex]{.4pt}{.77em}}}

\newtheorem{theorem}{Theorem}

\newcommand\norm[1]{\left\lVert#1\right\rVert}
\graphicspath{ {images} }
\definecolor{Gray}{gray}{0.85}

\title{Contrastive Loss is All You Need to Recover Analogies as Parallel Lines}

\definecolor{fl}{HTML}{00C8FF}

\definecolor{er}{HTML}{ff0000}

\author{Narutatsu Ri \\
  Columbia University \\
  \texttt{wl2787@columbia.edu} \\\And
  Fei-Tzin Lee \\
  Columbia University \\
  \texttt{feitzin@cs.columbia.edu} \\\And
  Nakul Verma \\
  Columbia University \\
  \texttt{verma@cs.columbia.edu} \\}

\begin{document}
\maketitle
\begin{abstract}

While static word embedding models are known to represent linguistic analogies as parallel lines in high-dimensional space, the underlying mechanism as to why they result in such geometric structures remains obscure. We find that an elementary contrastive-style optimization employed over distributional information performs competitively with popular word embedding models on analogy recovery tasks, while achieving dramatic speedups in training time. Further, we demonstrate that a contrastive loss is sufficient to create these parallel structures in word embeddings, and establish a precise relationship between the co-occurrence statistics and the geometric structure of the resulting word embeddings.
\end{abstract}

\section{Introduction}
Static word embeddings take inspiration from the distributional hypothesis \cite{Firth1957} and assign vector representations to words based on co-occurrence statistics. Such embeddings are known to implicitly encode syntactic and semantic analogies as parallelogram-type structures (\citealp{Mikolov2013EfficientEO, mikolov-etal-2013-linguistic}). This discovery inspired a series of theoretical investigations \cite{NIPS2014_feab05aa, gittens-etal-2017-skip, https://doi.org/10.48550/arxiv.1901.09813, ethayarajh-etal-2019-towards}. 

\begin{figure}[t!]
\centering
\includegraphics[width=0.50\textwidth]{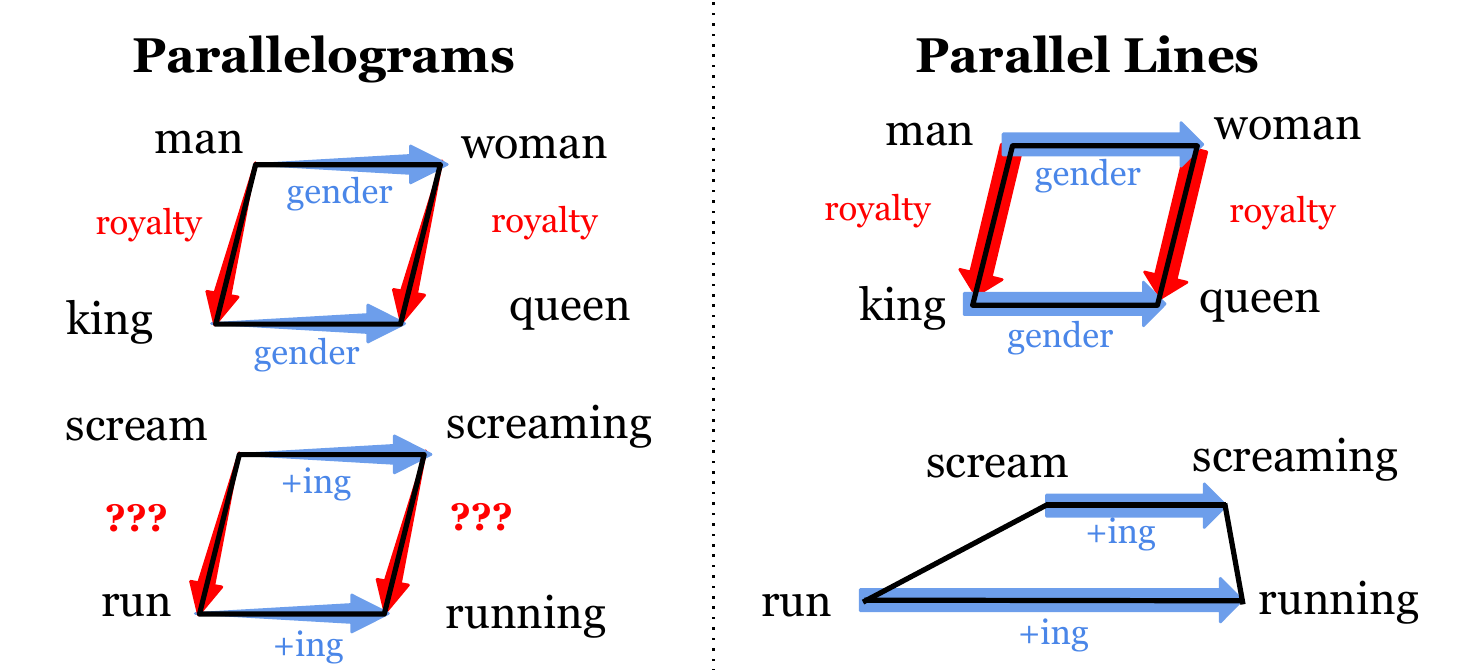}
\small{\caption{\label{fig:parallelograms_trapezoids_example} 
Visualization of analogies as parallelograms and as parallel lines. For the quadruple \texttt{``man, woman, king, queen"}, two analogy relations coincide (\texttt{``man:woman = king:queen"} representing gender and \texttt{``man:king = woman:queen"} representing royalty). In contrast, the quadruple \texttt{``run, running, scream screaming"} contains only one analogy relation (\texttt{``run:running = scream:screaming"} representing present participle). Representing analogies as lines relaxes the geometric requirements on the analogy structure.}}
\end{figure}

Recent studies reconsider whether analogies are indeed represented as parallelograms in the embedding space \cite{schluter-2018-word, linzen-2016-issues, https://doi.org/10.48550/arxiv.2102.11749}, and propose a weaker notion of viewing analogies as parallel \textit{lines} \cite{arora-etal-2016-latent} as a more appropriate model (cf.\ Figure \ref{fig:parallelograms_trapezoids_example}). While this claim is shown to hold empirically for popular word embeddings \cite{fournier-etal-2020-analogies}, few analyze the theoretical underpinnings of this phenomenon. 

In this paper, we present a remarkable observation that a simple contrastive-style optimization \cite{1467314} performs just as well as highly-optimized versions of popular word embeddings while achieving 50$\times$ speedup in training time. Our work theoretically analyzes the precise conditions under which this optimization procedure can recover analogies as parallel lines. We further investigate the extent to which real-world data satisfies these conditions, and the contrastive loss recovers such parallel structures.

In Section \ref{sec:related_work}, we review recent literature on the theory of word embeddings. Sections \ref{sec:cwm} and \ref{sec:analysis} present our contrastive learning objective and its analysis. Section \ref{sec:experiments} showcases the performance of our approach on analogy-based benchmarks.\footnote{Code can be found at \url{https://github.com/narutatsuri/cwm}.} 

\section{Related Work}\label{sec:related_work}
\textbf{Analogies as Parallelograms.} \citet{gittens-etal-2017-skip} study the parallelogram phenomenon by analyzing analogies as a relation between paraphrases. \citet{https://doi.org/10.48550/arxiv.1901.09813} extend this line of work and show that analogies are captured as parallelograms when the vectors are linear projections of the \emph{Pointwise Mutual Informaton} (PMI) matrix. \citet{ethayarajh-etal-2019-towards} further generalize Gittens' result by introducing the \textit{co-occurrence shifted pointwise mutual information} (csPMI)\footnote{csPMI$(a, b) = $ PMI$(a, b) + \log p(a, b)$.} and analyze the conditions on the csPMI for which parallelograms emerge. 

\textbf{Analogies as Parallel Lines.} To the best of our knowledge, the only theoretical work that explores analogies more generally as parallel lines is by \citet{arora2019latent}, who propose that analogies are encoded as such when the inner products between embeddings weakly recover the PMI of word co-occurrence statistics. We take an alternate approach and show that a contrastive-style optimization suffices to encode analogies as parallel lines.

\section{The Contrastive Word Model (CWM)\label{sec:cwm}}
Contrastive learning methods are based on an intuitive yet powerful idea that pulling similar items closer together while pushing dissimilar items away significantly improves model performance. 

We can employ the same push-pull dynamics in word embeddings by placing the vector representations of words that co-occur closer together than those of words that do not. We call this the Contrastive Word Model (CWM), detailed below.

\subsection{Notation \& Formulation}
Given a training corpus, we denote the vocabulary as $W$. We aim to learn a $D$-dimensional vector representation $v_w$ for each word $w$ in the vocabulary. The collection of all these vectors is denoted by $V = \{v_1,\ldots, v_{|W|}\}$. We refer to the length-normalized version of a vector ${v}$ as $\hat{v}$. 

Let $\#(i)$ be the occurrence count of word $i$ and $\#(i, j)$ the co-occurrence count (for a context window of size $\Delta$) of words $i$ and $j$ in the training corpus. We denote \emph{window words} as words that co-occur with a reference \emph{center word} (these are reminiscent of the target and context words in \citealp{mikolov-etal-2013-linguistic}), and \emph{negative window words} as words that do not co-occur with the center word. The center-, window-, and negative window words are denoted as $c, w, w'$ respectively. Let $D_{c, w}$ be the set of negative window words for fixed $c, w$. 
We define the CWM objective as:
\begin{align*}
\sum_{c \in W} \! \sum_{w \in W} \! \#(c, w) \cdot \!\!\!\!\!\!  \sum_{w' \in D_{c, w}} \!\!\!\! \Big[m - \underbrace{\hat{v}_c \cdot \hat{v}_w}_{\textup{pull}} + \underbrace{\hat{v}_c \cdot \hat{v}_{w'}}_{\textup{push}}\Big]_{_+},
\end{align*}
where $[\cdot]_{_+}$ is the hinge function and $m$ is a tunable hyperparameter. 

To better understand our proposed loss, consider its effect on a fixed center word $c$. The difference between the terms $\hat{v}_c \cdot \hat{v}_w$ and $\hat{v}_c \cdot \hat{v}_{w'}$ encourages the \emph{angle} between vectors $v_c$ and $v_w$ to be smaller than that between $v_c$ and $v_{w'}$ by at least a margin of $m$. The hinge function neutralizes the loss once the vectors satisfy the desired relationship. Such max-margin type losses among triples are well investigated in metric learning literature \cite{NIPS2005_a7f592ce}. 

\subsection{Relation to Popular Word Embeddings}
Interestingly, popular word embedding models such as Skip-gram \cite{Mikolov2013EfficientEO} and GloVe \cite{pennington-etal-2014-glove} can be viewed as implicitly employing a push-pull action similar to CWM. Consider Skip-gram's objective: for a fixed pair of co-occurring words $c$ and $w$, the model updates the word vector $v_c$ as:
\begin{align}\label{eq:sgns_update}
v_c^{\textup{new}} = & v_c^{\textup{old}} + \underbrace{\left(1-\frac{e^{v_w ^\intercal u_{c'}}}{\sum_{w' \in W} e^{v_w ^\intercal u_{w'}}} \right)}_{\textup{pull}}v_w\\
&- \underbrace{\mathbb{E}_{w'\sim W}[v_{w'}]}_{\textup{push}}\nonumber + \textup{additional terms}.
\end{align}
Here, the word $c'$ (and its target vector $u_{c'}$) co-occurs with both $c$ and $w$, encouraging all of them to be mapped together (pull), whereas the negative term pushes away randomly sampled words $w'$ from $c$. See Appendix \ref{app:sgns_update} for a derivation.

The GloVe objective, on the other hand, performs a series of updates on $c, w$, and $w'$ as:
\begin{align}\label{eq:glove_update}
\textup{pull } &
\Bigg\{ 
\begin{matrix}
v_c^{\textup{new}} &= v_c^{\textup{old}} + g(c, c')u_{c'}\\
v_w^{\textup{new}} &= v_w^{\textup{old}} + g(w, c')u_{c'}
\end{matrix} \\
\textup{push } &
\Big\{ v_{w'}^{\textup{new}} \;\;\;= v_{w'}^{\textup{old}} - g(w', c')u_{c'}, \nonumber
\end{align}
where $g(\cdot, \cdot)$ always returns a positive value. Notice that the \emph{positive} contribution of $g$ in the first two updates encourages $v_c$ and $v_w$ to be closer together (pull), while the \emph{negative} contribution to the $v_{w'}$ update encourages it to be pushed away. See Appendix \ref{app:glove_update} for a derivation.

We believe that part of the success of these word embedding models is due to their implicit push-pull dynamics. Hence, a natural question to consider is what happens when one purely optimizes for the push-pull action alone.

\section{Analysis\label{sec:analysis}}
In this section, we provide a theoretical justification for the emergence of analogies as parallel lines when we optimize for the CWM objective. 

Consider the expression for word vectors $v_c\in V$ that minimizes the global objective:
\begin{align} \label{eq:optimal_embeddings}
\!\!\!\!{v}_c = \rho_c \Bigg( \! \sum_{w \in W} \!\! \Big(\frac{\#(c, w)}{\#(c)} {\hat{v}}_w\Big) - \!\!\!\! \underset{w'\sim U(W)}{\mathbb{E}} \!\! \left[\hat{v}_{w'}\right] \Bigg),
\end{align}
where $\rho_c \in \mathbb{R}$ is a constant dependent on $c$. 
In essence, $v_c$ is the difference between the weighted average of the window words and the mean of all word vectors. See Appendix \ref{app:superposition} for derivation.

Under Eq.\ \eqref{eq:optimal_embeddings}, we consider the conditions that word co-occurrence statistics need to satisfy for a set of words $a, b, c, d$ to form parallel geometric structures. 

\begin{theorem}\label{thm:main_theorem}
For any quadruple of words $a, b, c, d \in W$, if there exists a constant $\zeta \in \mathbb{R}$ where the co-occurrence statistics satisfy the condition: $\quad\forall w \in W$
\begin{equation}\label{eq:2}
\!\!\!\!\resizebox{.9\hsize}{!}{$
\left(\frac{\#(a, w)}{\#(a)} - \frac{\#(b, w)}{\#(b)}\right)\!\!\!\bigg/\!\!\!\left(\frac{\#(c, w)}{\#(c)} - \frac{\#(d, w)}{\#(d)}\right) := \zeta,$}
\end{equation}
then the corresponding word vectors satisfy the property:
$$
{\hat{v}}_a - {\hat{v}}_b = \zeta\left({\hat{v}}_c - {\hat{v}}_d \right).
$$
\end{theorem}
Note that Theorem \ref{thm:main_theorem} establishes a direct relationship between word co-occurrence statistics---which are solely derived from the training corpus---and the geometric structure of the word embedding. 

For a given quadruple $a, b, c, d\in W$ (regardless of whether they form an analogy), the existence of $\zeta$ induces parallel structures between $\hat{v}_a, \hat{v}_b, \hat{v}_c, \hat{v}_d$. 
If such a $\zeta$ exists and is equal to $1$, then $\hat{v}_b - \hat{v}_a = \hat{v}_d - \hat{v}_c$ and the quadruple forms a parallelogram. When $\zeta \neq 1$, then the difference vectors $\hat{v}_b - \hat{v}_a$ and $\hat{v}_d - \hat{v}_c$ are mainly parallel, inducing a trapezoidal structure among $\hat{v}_a, \hat{v}_b, \hat{v}_c, \hat{v}_d$ (cf.\ Figure \ref{fig:parallelograms_trapezoids_example}). 

One would expect that the co-occurrence statistics of real data conform with the existence of such a $\zeta$ value for analogy quadruples, whereas $\zeta$ does not exist for random quadruples. This is empirically investigated in Section \ref{sec:analogies_and_lines}. The relationship between the value of $\zeta$ and the resulting parallelogram structure (parallelogram vs.\ trapezoid) is empirically verified in Section \ref{sec:zeta_parallelograms}. 

\section{Experiments}\label{sec:experiments}

We first compare the performance of CWM to that of other popular word embedding methods on analogy recovery (Section \ref{sec:results-analogy}). We then empirically verify the degree to which our assumptions regarding co-occurrences hold on real data (Section \ref{sec:analogies_and_lines}) as well as the relation between $\zeta$ and the parallelogram structure (Section \ref{sec:zeta_parallelograms}).

\subsection{Data and Training Procedure}
We use the 03/2023 version of Wikimedia Downloads dump \cite{wikidump} and train CWM for a single pass over the corpus using $\Delta = 5$ and $m=0.2$ (chosen via cross validation from the range $0.1\sim1$). For comparison, we also train Skip-gram with Negative Sampling (SGNS) and GloVe over the same corpus with the default parameter settings provided by \citet{mikolov-etal-2013-linguistic} and \citet{pennington-etal-2014-glove} respectively.

We utilize the BATS analogy dataset \citep{gladkova-etal-2016-analogy} for all analogy related tasks. For all word embeddings, we use dimension $D=300$ and the vectors are length-normalized to follow practical conventions \cite{mikolov-etal-2013-linguistic}. Training was done on 256 instances of AMD EPYC 7763 64-Core Processor machine.

\begin{table}[t]
\centering
\begin{tabular}{l|cc|cc}
\hline
& \multicolumn{2}{|c|}{Analogies} & \multicolumn{2}{c}{Training}\\
\hline
\textbf{Model} & \textbf{PCS} & \textbf{MSM} & \textbf{Time (hrs)} & \textbf{Speedup}\\
\hline
\rowcolor{Gray}
CWM & \textbf{0.677} & \textbf{0.469}& \textbf{0.59} & \textbf{49$\times$}\\
SGNS & 0.675 & 0.433 & 29.27 & 1$\times$\\
GloVe & 0.667 & 0.423 & 30.71 & 0.91$\times$\\
\hline
\end{tabular}
\small{\caption{\label{table:metric_performances} Performances for word embedding models. CWM refers to our contrastive word model. SGNS refers to Skip-gram with negative sampling. Best numbers are bolded.}}
\end{table}

\subsection{Analogy Recovery}\label{sec:results-analogy}
To assess the degree to which word embeddings encode analogies as lines consistently in the embedding space, we use two intuitive metrics proposed by 
\citet{fournier-etal-2020-analogies}:
the \textit{Pairing Consistency Score} (PCS) and \textit{Mean Similarity Measure} (MSM). PCS assesses analogy alignment precision (the number of non-analogy offsets incorrectly aligned with true analogy offsets), while MSM measures absolute alignment. 

Table \ref{table:metric_performances} shows the relative performance of popular word embeddings. Notice that our method performs $7\%$ better than popular word embeddings on the MSM metric, indicating that the word vectors learned by CWM exhibit higher alignment among analogy quadruples than Skip-gram and GloVe. CWM's performance on the PCS metric indicates that parallel lines are not erroneously encoded for non-analogy words. For completeness, see Appendix \ref{app:analogy_parallelograms} for parallelogram recovery performances (previous literature questions the validity of the standard evaluation method).

\begin{figure}[t]
    \vspace{-1em}
    \centering
    \includegraphics[width=0.50\textwidth]{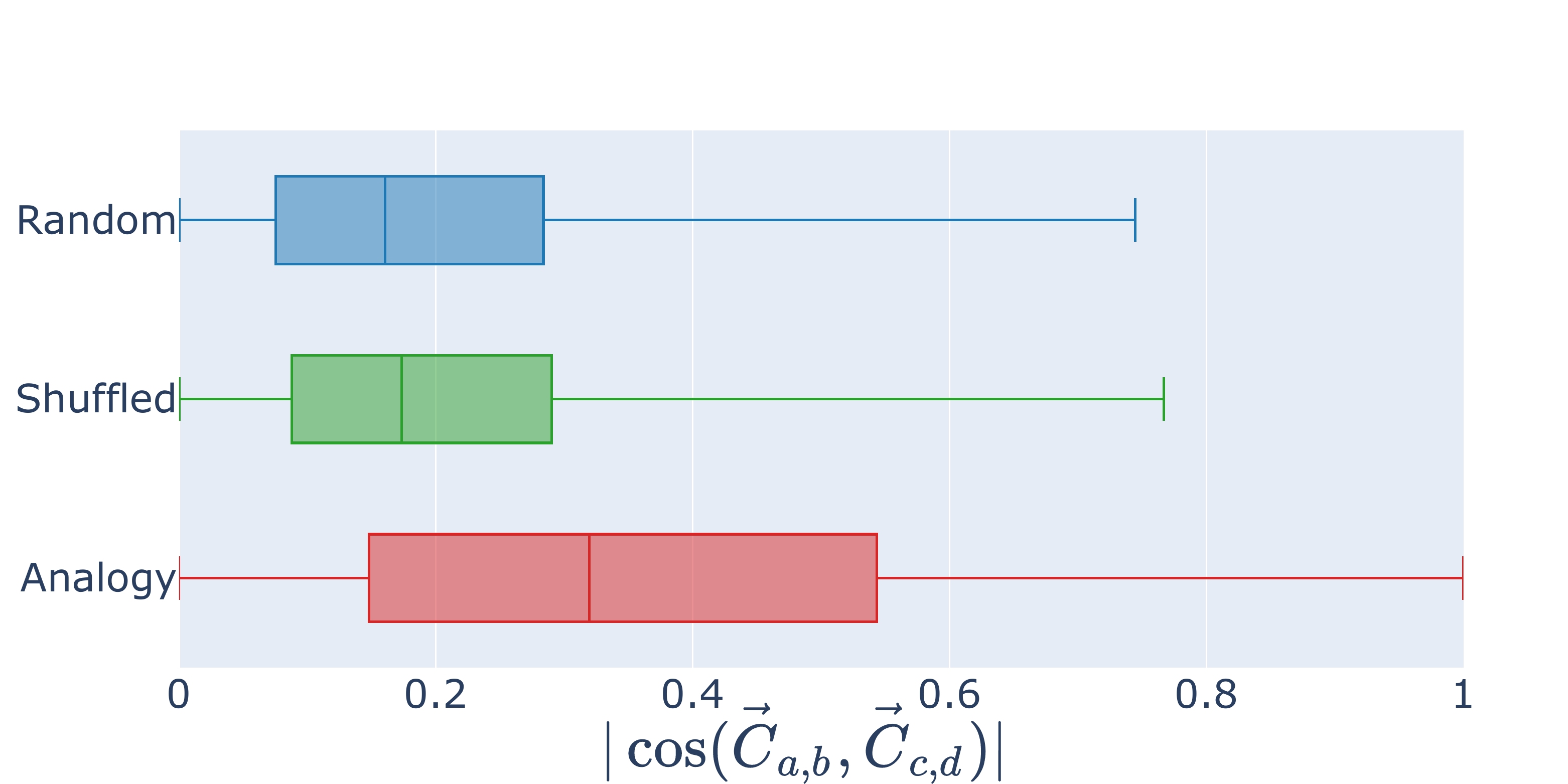}
    \small{\caption{\label{fig:zeta_concentration} Cosine similarities between co-occurrence vectors $\vec{C}_{a, b}$ and $\vec{C}_{c, d}$ for words $a, b, c, d$ from uniformly sampled word quadruples (Random), shuffled analogy pairs (Shuffled), and true analogy pairs (Analogy).}}
\end{figure}

\subsection{Existence of $\zeta$ and Analogies\label{sec:analogies_and_lines}}
Theorem \ref{thm:main_theorem} provides insight into the conditions required for CWM to induce parallel lines in the learned word vectors, but these conditions are not specific to analogy word pairs. Thus, the question remains: does $\zeta$ exist only when a quadruple forms an analogy?

Here, we study the level at which the co-occurrence statistics of analogy and non-analogy pairs satisfy the condition in Theorem \ref{thm:main_theorem}. To assess the existence of $\zeta$, consider the vectors $\vec{C}_{a, b}, \vec{C}_{c, d} \in \mathbb{R}^{|W|}$ (derived purely from co-occurrence counts): 
\begin{align*}
\resizebox{.9\hsize}{!}{$
\vec{C}_{a, b} = \Bigg[\left(\frac{\#(a, w_1)}{\#(a)} - \frac{\#(b, w_1)}{\#(b)}\right), \dots, \left(\frac{\#(a, w_{|W|})}{\#(a)} - \frac{\#(b, w_{|W|})}{\#(b)}\right)\Bigg]$},\\
\resizebox{.9\hsize}{!}{$
\vec{C}_{c, d} = \Bigg[\left(\frac{\#(c, w_1)}{\#(c)} - \frac{\#(d, w_1)}{\#(d)}\right), \dots, \left(\frac{\#(c, w_{|W|})}{\#(c)} - \frac{\#(d, w_{|W|})}{\#(d)}\right)\Bigg]$}.
\end{align*}
Existence of a $\zeta$ where Eq.\ \eqref{eq:2} holds for $a, b, c, d$, implies that all entries in $\vec{C}_{a, b}$ are equal to the corresponding entries in $\vec{C}_{c, d}$ scaled by a factor of $\zeta$. This indicates that when $\zeta$ exists, $\vec{C}_{a, b}$ and $\vec{C}_{c, d}$ are collinear. Thus, we can approximate assessing the existence of $\zeta$ by evaluating whether the cosine similarity between $\vec{C}_{a, b}$ and $\vec{C}_{c, d}$ is sufficiently high. 

We consider three settings from which the quadruples are obtained: randomly sampled word quadruples, false shuffled analogies, and true analogies using the BATS dataset. We compute the distribution of cosine similarities for all quadruples from these settings. 

Results are shown in Figure \ref{fig:zeta_concentration}. Observe that the cosine similarities of random and shuffled quadruples is significantly lower than that for analogy words. This indicates a positive association between $\zeta$ and analogy word quadruples in real world corpora. 



\begin{table}[t]
\centering
\scalebox{0.9}{
\begin{tabular}{l|c}
\hline 
\textbf{Analogy Quadruple} & \textbf{Sim.}\\
\hline
{\small\texttt{fall:rise = under:over}} & 1.000\\
{\small\texttt{prevent:preventing = follow:following}} & 0.9901 \\
{\small\texttt{lancaster:lancashire = salford:manchester}} & 0.9812\\
{\small\texttt{refer:referred = agree:agreed}} & 0.9740\\
\hline
{\small\texttt{organized:arranged = dollars:bucks}} & 0.0006\\
{\small\texttt{staircase:step = shilling:pence}} & 0.0006\\
{\small\texttt{guitar:string = church:altar}} & 0.0004\\
{\small\texttt{monkey:infant = fox:cub}} & 0.0001\\
\hline
\end{tabular}}
\small{\caption{\label{table:good_and_bad_analogies} Samples of analogy quadruples illustrating cosine similarity values between $\vec{C}_{a,b}$ and $\vec{C}_{c,d}$. "Sim." denotes the value of $|\!\cos(\vec{C}_{a,b}, \vec{C}_{c,d})|$.}}
\end{table}

Furthermore, it is worth noting the presence of "ambiguous" analogies within the BATS dataset. These include
analogies with valid alternative replacements (e.g.\ \texttt{sun:orange = sea:blue} can also be \texttt{sun:red = sea:blue}), or analogies with unclear relationships (e.g.\ lexicographic analogies such as \texttt{father:dad = lady:madam}). We investigate whether the ambiguity of an analogy correlates with its cosine similarity between $\vec{C}_{a,b}$ and $\vec{C}_{c,d}$
by sampling from analogy quadruples with high and low values of $|\!\cos(\vec{C}_{a,b}, \vec{C}_{c,d})|$. 

Results are shown in Table \ref{table:good_and_bad_analogies}. Observe that analogy quadruples with high cosine similarity between $\vec{C}_{a,b}$ and $\vec{C}_{c,d}$ seems to demonstrate a clear relationships, whereas those with low cosine similarity exhibit weaker/ambiguous relationships. 



\subsection{$\zeta$ and Geometric Structure\label{sec:zeta_parallelograms}}

We now examine the effect of $\zeta$ on the geometry of analogy word pairs. Recall that $\zeta$ exists for quadruples where $|\!\cos(\vec{C}_{a, b}, \vec{C}_{c, d})| = 1$. As this condition is unlikely to hold exactly on real data, we approximate $\zeta$ with the ratio $\hat \zeta := \|\vec{C}_{a, b}\|/\|\vec{C}_{c, d}\|$ for quadruples with high cosine similarity (which we define as $|\!\cos(\vec{C}_{a, b}, \vec{C}_{c, d})|\geq0.9$). We expect the word vectors to form parallelograms when $\hat \zeta \approx 1$ ($0.95\leq\hat\zeta\leq1.05$), and form trapezoids otherwise. 

Specifically, for each such quadruple, we compute the word $w$ that minimizes $\|\hat{v}_b - \hat{v}_a + \hat{v}_c - \hat{v}_w\|$ for parallelograms; ideally, $w$ should equal $d$. For trapezoids, we retrieve the word $w$ that maximizes the quantity $\cos(\hat{v}_b-\hat{v}_a, \hat{v}_w-\hat{v}_c)$. If the word $d$ is among the top $k$ words, we deem the quadruple to satisfy the corresponding geometric structure. For both cases, we consider $k = 1$ and $5$. 

\begin{table}[t]
\centering
\begin{tabular}{c|cc}
\hline
 & $k=1$  & $k=5$\\
\hline \\[-1em]
$\hat \zeta \not\approx 1$ & 0.800 \small{(619/774)} & 0.862 \small{(667/774)} \\
\hline \\[-1em]
$\hat \zeta \approx 1$ & 0.652 \small{(137/210)} & 0.871 \small{(183/210)} \\
\hline
\end{tabular}
\small{\caption{\label{table:zeta_recovery}Parallelogram/trapezoid recovery performances for different values of $\hat\zeta$. Parallelogram recovery for all analogy pairs is $0.27$ (see Table \ref{table:performances_sup} in Appendix), indicating dramatic performance increase for the analogy subset where $\hat\zeta \approx 1$.}}
\end{table}

Results are shown in Table \ref{table:zeta_recovery}. Observe that for $k=5$, $87\%$ of the quadruples form parallelograms when $\hat \zeta \approx 1$ (i.e.,\ $0.95\leq\hat\zeta\leq1.05$), and $86\%$ of quadruples form trapezoid-type structures when $\hat\zeta \not\approx 1$. This validates our expectation that parallelograms and trapezoids indeed form when $\hat\zeta \approx 1$ and $ \hat\zeta \not\approx 1$ respectively.

\section{Conclusion and Discussion}\label{sec:conclusion}
We demonstrate that optimizing a contrastive-style objective over word co-occurrences is indeed sufficient to encode analogies as parallel lines. Our analysis (Theorem \ref{thm:main_theorem}) sheds light on the inner workings of word embeddings:\ parallel geometry is induced largely from word co-occurrence statistics for any push-pull model. Our work builds upon and generalizes previous literature that illuminates the underlying mechanisms governing the geometry of word embeddings. 

Note that while our results demonstrate the sufficiency of the push-pull mechanism for recovering analogies as parallel lines, it remains unclear whether push-pull is a necessary condition for this phenomenon. Investigating alternative mechanisms and their ability to achieve similar results would provide further insight into the relationship between word co-occurrence statistics and the recovery of analogies. 

\bibliography{custom}

\begin{thebibliography}{21}
\expandafter\ifx\csname natexlab\endcsname\relax\def\natexlab#1{#1}\fi

\bibitem[{Allen and
  Hospedales(2019)}]{https://doi.org/10.48550/arxiv.1901.09813}
Carl Allen and Timothy Hospedales. 2019.
\newblock \href {https://doi.org/10.48550/ARXIV.1901.09813} {Analogies
  explained: Towards understanding word embeddings}.

\bibitem[{Arora et~al.(2016)Arora, Li, Liang, Ma, and
  Risteski}]{arora-etal-2016-latent}
Sanjeev Arora, Yuanzhi Li, Yingyu Liang, Tengyu Ma, and Andrej Risteski. 2016.
\newblock \href {https://doi.org/10.1162/tacl_a_00106} {A latent variable model
  approach to {PMI}-based word embeddings}.
\newblock \emph{Transactions of the Association for Computational Linguistics},
  4:385--399.

\bibitem[{Arora et~al.(2019)Arora, Li, Liang, Ma, and
  Risteski}]{arora2019latent}
Sanjeev Arora, Yuanzhi Li, Yingyu Liang, Tengyu Ma, and Andrej Risteski. 2019.
\newblock \href {http://arxiv.org/abs/1502.03520} {A latent variable model
  approach to pmi-based word embeddings}.

\bibitem[{Bruni et~al.(2014)Bruni, Tran, and Baroni}]{Bruni2014MultimodalDS}
Elia Bruni, Nam~Khanh Tran, and Marco Baroni. 2014.
\newblock Multimodal distributional semantics.
\newblock \emph{J. Artif. Intell. Res.}, 49:1--47.

\bibitem[{Chopra et~al.(2005)Chopra, Hadsell, and LeCun}]{1467314}
S.~Chopra, R.~Hadsell, and Y.~LeCun. 2005.
\newblock \href {https://doi.org/10.1109/CVPR.2005.202} {Learning a similarity
  metric discriminatively, with application to face verification}.
\newblock In \emph{2005 IEEE Computer Society Conference on Computer Vision and
  Pattern Recognition (CVPR'05)}, volume~1, pages 539--546 vol. 1.

\bibitem[{Ethayarajh et~al.(2019)Ethayarajh, Duvenaud, and
  Hirst}]{ethayarajh-etal-2019-towards}
Kawin Ethayarajh, David Duvenaud, and Graeme Hirst. 2019.
\newblock \href {https://doi.org/10.18653/v1/P19-1315} {Towards understanding
  linear word analogies}.
\newblock In \emph{Proceedings of the 57th Annual Meeting of the Association
  for Computational Linguistics}, pages 3253--3262, Florence, Italy.
  Association for Computational Linguistics.

\bibitem[{Finkelstein et~al.(2002)Finkelstein, Gabrilovich, Matias, Rivlin,
  Solan, Wolfman, and Ruppin}]{Finkelstein2002PlacingSI}
Lev Finkelstein, Evgeniy Gabrilovich, Y.~Matias, Ehud Rivlin, Zach Solan, Gadi
  Wolfman, and Eytan Ruppin. 2002.
\newblock Placing search in context: the concept revisited.
\newblock \emph{ACM Trans. Inf. Syst.}, 20:116--131.

\bibitem[{Firth(1957)}]{Firth1957}
J.~Firth. 1957.
\newblock A synopsis of linguistic theory 1930-1955.
\newblock In \emph{Studies in Linguistic Analysis}. Philological Society,
  Oxford.
\newblock Reprinted in Palmer, F. (ed. 1968) Selected Papers of J. R. Firth,
  Longman, Harlow.

\bibitem[{Foundation(2023)}]{wikidump}
Wikimedia Foundation. 2023.
\newblock \href {https://dumps.wikimedia.org} {Wikimedia downloads}.

\bibitem[{Fournier and
  Dunbar(2021)}]{https://doi.org/10.48550/arxiv.2102.11749}
Louis Fournier and Ewan Dunbar. 2021.
\newblock \href {https://doi.org/10.48550/ARXIV.2102.11749} {Paraphrases do not
  explain word analogies}.

\bibitem[{Fournier et~al.(2020)Fournier, Dupoux, and
  Dunbar}]{fournier-etal-2020-analogies}
Louis Fournier, Emmanuel Dupoux, and Ewan Dunbar. 2020.
\newblock \href {https://doi.org/10.18653/v1/2020.conll-1.29} {Analogies minus
  analogy test: measuring regularities in word embeddings}.
\newblock In \emph{Proceedings of the 24th Conference on Computational Natural
  Language Learning}, pages 365--375, Online. Association for Computational
  Linguistics.

\bibitem[{Gittens et~al.(2017)Gittens, Achlioptas, and
  Mahoney}]{gittens-etal-2017-skip}
Alex Gittens, Dimitris Achlioptas, and Michael~W. Mahoney. 2017.
\newblock \href {https://doi.org/10.18653/v1/P17-1007} {Skip-gram - {Z}ipf +
  uniform = vector additivity}.
\newblock In \emph{Proceedings of the 55th Annual Meeting of the Association
  for Computational Linguistics (Volume 1: Long Papers)}, pages 69--76,
  Vancouver, Canada. Association for Computational Linguistics.

\bibitem[{Gladkova et~al.(2016)Gladkova, Drozd, and
  Matsuoka}]{gladkova-etal-2016-analogy}
Anna Gladkova, Aleksandr Drozd, and Satoshi Matsuoka. 2016.
\newblock \href {https://doi.org/10.18653/v1/N16-2002} {Analogy-based detection
  of morphological and semantic relations with word embeddings: what works and
  what doesn{'}t.}
\newblock In \emph{Proceedings of the {NAACL} Student Research Workshop}, pages
  8--15, San Diego, California. Association for Computational Linguistics.

\bibitem[{Hill et~al.(2015)Hill, Reichart, and
  Korhonen}]{hill-etal-2015-simlex}
Felix Hill, Roi Reichart, and Anna Korhonen. 2015.
\newblock \href {https://doi.org/10.1162/COLI_a_00237} {{S}im{L}ex-999:
  Evaluating semantic models with (genuine) similarity estimation}.
\newblock \emph{Computational Linguistics}, 41(4):665--695.

\bibitem[{Levy and Goldberg(2014)}]{NIPS2014_feab05aa}
Omer Levy and Yoav Goldberg. 2014.
\newblock \href
  {https://proceedings.neurips.cc/paper/2014/file/feab05aa91085b7a8012516bc3533958-Paper.pdf}
  {Neural word embedding as implicit matrix factorization}.
\newblock In \emph{Advances in Neural Information Processing Systems},
  volume~27. Curran Associates, Inc.

\bibitem[{Linzen(2016)}]{linzen-2016-issues}
Tal Linzen. 2016.
\newblock \href {https://doi.org/10.18653/v1/W16-2503} {Issues in evaluating
  semantic spaces using word analogies}.
\newblock In \emph{Proceedings of the 1st Workshop on Evaluating Vector-Space
  Representations for {NLP}}, pages 13--18, Berlin, Germany. Association for
  Computational Linguistics.

\bibitem[{Mikolov et~al.(2013{\natexlab{a}})Mikolov, Chen, Corrado, and
  Dean}]{Mikolov2013EfficientEO}
Tomas Mikolov, Kai Chen, Gregory~S. Corrado, and Jeffrey Dean.
  2013{\natexlab{a}}.
\newblock Efficient estimation of word representations in vector space.
\newblock In \emph{International Conference on Learning Representations}.

\bibitem[{Mikolov et~al.(2013{\natexlab{b}})Mikolov, Yih, and
  Zweig}]{mikolov-etal-2013-linguistic}
Tomas Mikolov, Wen-tau Yih, and Geoffrey Zweig. 2013{\natexlab{b}}.
\newblock \href {https://aclanthology.org/N13-1090} {Linguistic regularities in
  continuous space word representations}.
\newblock In \emph{Proceedings of the 2013 Conference of the North {A}merican
  Chapter of the Association for Computational Linguistics: Human Language
  Technologies}, pages 746--751, Atlanta, Georgia. Association for
  Computational Linguistics.

\bibitem[{Pennington et~al.(2014)Pennington, Socher, and
  Manning}]{pennington-etal-2014-glove}
Jeffrey Pennington, Richard Socher, and Christopher Manning. 2014.
\newblock \href {https://doi.org/10.3115/v1/D14-1162} {{G}lo{V}e: Global
  vectors for word representation}.
\newblock In \emph{Proceedings of the 2014 Conference on Empirical Methods in
  Natural Language Processing ({EMNLP})}, pages 1532--1543, Doha, Qatar.
  Association for Computational Linguistics.

\bibitem[{Schluter(2018)}]{schluter-2018-word}
Natalie Schluter. 2018.
\newblock \href {https://doi.org/10.18653/v1/N18-2039} {The word analogy
  testing caveat}.
\newblock In \emph{Proceedings of the 2018 Conference of the North {A}merican
  Chapter of the Association for Computational Linguistics: Human Language
  Technologies, Volume 2 (Short Papers)}, pages 242--246, New Orleans,
  Louisiana. Association for Computational Linguistics.

\bibitem[{Weinberger et~al.(2005)Weinberger, Blitzer, and
  Saul}]{NIPS2005_a7f592ce}
Kilian~Q Weinberger, John Blitzer, and Lawrence Saul. 2005.
\newblock \href
  {https://proceedings.neurips.cc/paper_files/paper/2005/file/a7f592cef8b130a6967a90617db5681b-Paper.pdf}
  {Distance metric learning for large margin nearest neighbor classification}.
\newblock In \emph{Advances in Neural Information Processing Systems},
  volume~18. MIT Press.

\end{thebibliography}
\bibliographystyle{acl_natbib}

\clearpage
\newpage

\appendix
\label{appendix}
\section{Proofs}
\subsection{Derivation of Eq.\ \eqref{eq:optimal_embeddings}}\label{app:superposition}

Recall that the global objective of CWM for the vocabulary $W$ and set of word vectors $V$ can be written as:
\begin{align*}
& \mathcal{L}(V) = \sum_{c \in W} \sum_{w \in W} \#(c, w) \\
& \sum_{w' \in D_{c, w}} \Big[m -  \hat{v}_c \cdot \hat{v}_w +  \hat{v}_c \cdot \hat{v}_{w'} \Big]_{_+},
\end{align*}
where $D_{c, w} = \{w' | w' \sim U(W)\}, |D_{c, w}| = k$ denotes the set of $k$ negative window words sampled uniformly from the vocabulary for each $c, w$ word pair and $U(W)$ denotes the uniform distribution over the vocabulary.

For fixed $c, w, w'$, consider the two cases where $m -  \hat{v}_c \cdot \hat{v}_w +  \hat{v}_c \cdot \hat{v}_{w'} > 0$ and $m - \hat{v}_c \cdot \hat{v}_w +  \hat{v}_c \cdot \hat{v}_{w'} \leq 0$. As the word vectors are not updated for the latter case, we examine the former by taking the partial derivative of $\mathcal{L}(V)$ with respect to ${v}_c$ and setting it to ${0}$:
\begin{align*}
&{0} = -\sum_{w \in W}\#(c, w) \\
& \sum_{w'\in D_{c, w}} \left( \frac{{v}_w}{\|{v}_c\| \|{v}_w\|} - \frac{{v}_{w'}}{\|{v}_c\| \|{v}_{w'}\|}\right. \\
& \left. \vphantom{\frac{{v}_w}{\|{v}_c\| \|{v}_w\|} - \frac{{v}_{w'}}{\|{v}_c\|  \|{v}_{w'}\|}}  + \left(  \frac{\hat{v}_c \cdot \hat{v}_{w'}}{\|{v}_c\|^2} -  \frac{(\hat{v}_c \cdot \hat{v}_w)}{\|{v}_c\|^2} \right) {v}_c \right)\\
\Leftrightarrow & \sum_{w \in W}\#(c, w) \sum_{w'\in D_{c, w}} \frac{{v}_w}{\|{v}_c\| \|{v}_w\|} \\
& -\sum_{w \in W}\#(c, w) \sum_{w'\in D_{c, w}} \frac{{v}_{w'}}{\|{v}_c\| \|{v}_{w'}\|} \\
= & \sum_{w\in W} \#(c, w) \sum_{w' \in D_{c, w}}\left( \frac{{v}_c  {v}_w}{\|{v}_c\|^2  \|{v}_w\|} \right.\\
& \left. - \frac{{v}_c  {v}_{w'}}{\|{v}_c\|^2  \|{v}_{w'}\|} \right) \frac{{v}_c}{\|{v}_c\|}.
\end{align*}
As $\sum_{w \in W}\#(c, w) \sum_{w'\in D_{c, w}} \frac{{v}_{w'}}{\|{v}_{w'}\|}$ represents $\sum_{w \in W} \#(c, w) \cdot k = k \cdot \#(c)$ uniform i.i.d. draws from the vocabulary, the following holds for sufficiently large values of $k\cdot \#(c)$:
\begin{align*}
\sum_{w \in W}\#(c, w) &\sum_{w'\in D_{c, w}} \frac{{v}_{w'}}{\|{v}_{w'}\|} =\\
&k \#(c) \mathbb{E}_{w' \sim U(W)} \left[ \frac{{v}_{w'}}{\|{v}_{w'}\|} \right].
\end{align*}

Setting $\mathbb{E}_{w' \sim U(W)} \left[ \frac{{v}_{w'}}{\|{v}_{w'}\|} \right] = {v}_p$ and dividing both sizes by $\frac{k\#(c)}{\|v\|}$, 
\begin{align*}
& \sum_{w\in W} \frac{\#(c, w)}{\#(c)} \frac{{v}_w}{\|{v}_w\|} - {v}_p\\
& = \left[\frac{{v}_c}{\|{v}_c\|} \left( \sum_{w\in W} \frac{\#(c, w)}{\#(c)} \frac{{v}_w}{\|{v}_w\|} - {v}_p \right) \right] \odot \frac{{v}_c}{\|{v}_c\|}.
\end{align*}
Setting $\sum_{w\in W} \frac{\#(c, w)}{\#(c)} \frac{{v}_w}{\|{v}_w\|} = {v}_{p'}$ and 
$
\gamma_c = \norm{\frac{{v}_c}{\|{v}_c\|} \left( \sum_{w\in W} \frac{\#(c, w)}{\#(c)} \frac{{v}_w}{\|{v}_w\|} - {v}_p \right)},
$
the above equation can be rewritten as:
$$
\frac{{v}_c}{\|{v}_c\|} = \frac{{v}_{p'}}{\gamma_c} \cdot \frac{1}{\norm {\frac{{v}_c}{\|{v}_c\|} }} \cdot \frac{1}{\cos \theta}.
$$
where $\theta$ indicates the angle between ${v}_{p'}$ and $\frac{{v}_c}{\|{v}_c\|}$. 

As $\norm {\frac{{v}_c}{\|{v}_c\|}} = 1$, 
$$
{v}_c = \|{v}_c\| \cdot \frac{{v}_{p'}}{\gamma_c}\cdot \frac{1}{\cos \theta}
$$
Notice that ${v}_c \parallel {v}_{p'}$ by the above construction , so $\cos\theta = 1$. Thus, 
$$
{v}_c = \|{v}_c\| \cdot \frac{{v}_{p'}}{\gamma_c} = \frac{\alpha\#(c)^{\frac{1}{\beta}}}{{\gamma_c}} \cdot {v}_{p'}. \quad \blacksquare
$$

The second equality is derived from the empirically observed property $\|{v}_c\| \propto \#(c)^\frac{1}{\beta}$ for some constant $\beta \in \mathbb{R}$, which is verified below. 

\begin{figure}[h]
\centering
\includegraphics[width=0.45\textwidth]{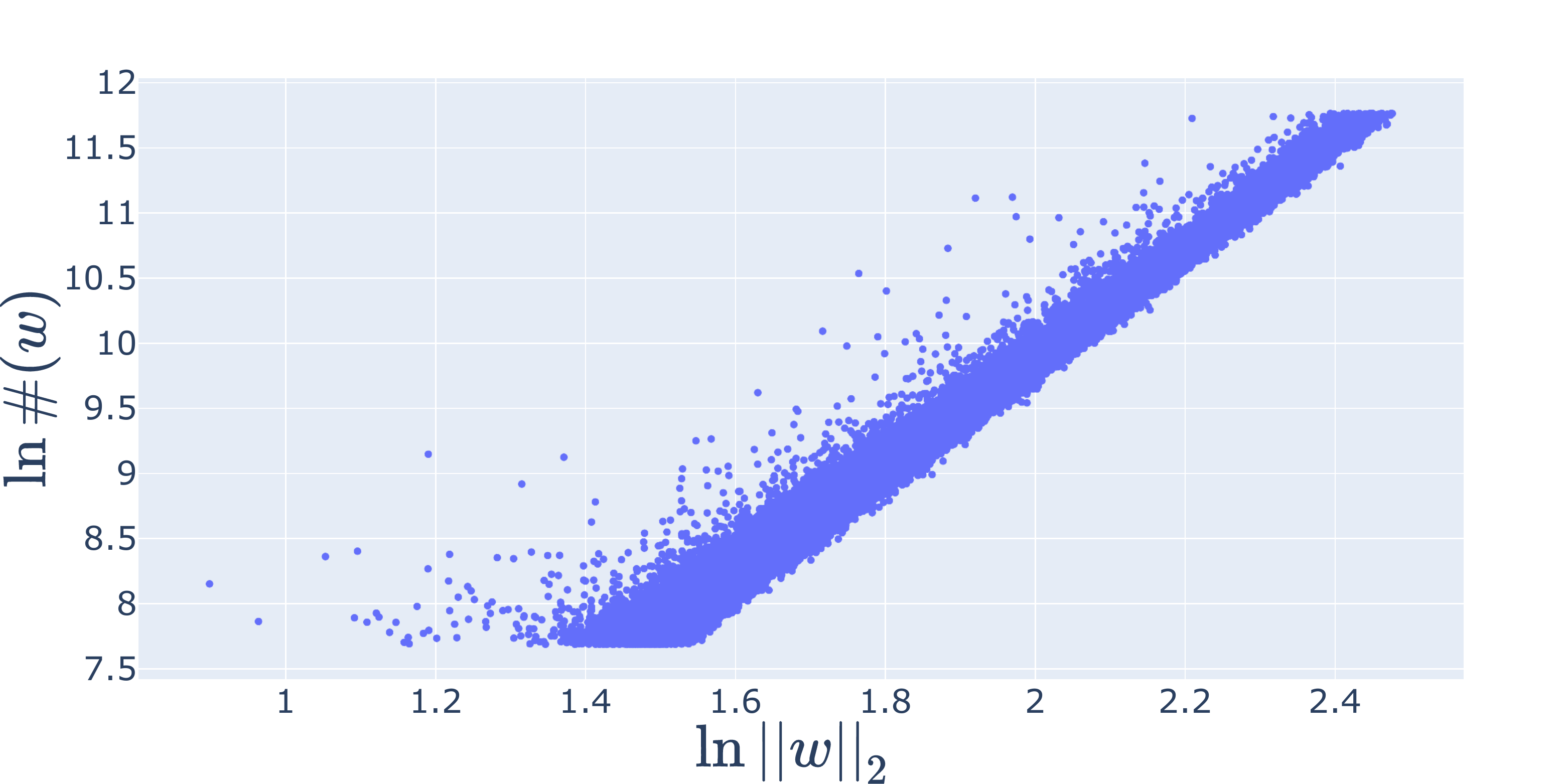}
\end{figure}

Interestingly, a similar linear relationship is also observed in existing word embedding models \cite{arora-etal-2016-latent}. 

\subsection{Proof for Theorem \ref{thm:main_theorem}}\label{app:zeta_cooccurrence}
Under the assumption that Eq.\ \eqref{eq:optimal_embeddings} holds, we can write the expressions for ${\hat{v}}_a - {\hat{v}}_b, {\hat{v}}_c - {\hat{v}}_d$ as follows.
\begin{align*}
{\hat{v}}_a - {\hat{v}}_b & = \frac{1}{\gamma_a}\left( \sum_{w \in W} \left(\frac{\#(a, w)}{\#(a)} \frac{{v}_w}{\|{v}_w\|}\right) - {v}_p \right) \\
& - \frac{1}{\gamma_b}\left( \sum_{w \in W} \left(\frac{\#(b, w)}{\#(b)} \frac{{v}_w}{\|{v}_w\|}\right) - {v}_p \right).
\end{align*} 

Under the assumption that $\forall c \in W: \gamma_c = \gamma$ for some $\gamma \in \mathbb{R}$, 
\begin{align*}
{\hat{v}}_a - {\hat{v}}_b = \frac{1}{\gamma} \sum_{w \in W} \left( \frac{\#(a, w)}{\#(a)} -\frac{\#(b, w)}{\#(b)} \right)\frac{{v}_w}{\|{v}_w\|}
\end{align*}

Using Eq.\ \eqref{eq:2}, 
\begin{align*}
{\hat{v}}_a - {\hat{v}}_b &= \frac{\zeta}{\gamma} \sum_{w \in W} \left( \frac{\#(c, w)}{\#(c)} -\frac{\#(d, w)}{\#(d)} \right)\frac{{v}_w}{\|{v}_w\|}\\
&= \zeta({\hat{v}}_c - {\hat{v}}_d). \quad\blacksquare
\end{align*}

The invariance of the value $\gamma_c$ can be verified through randomly sampling 5000 words $c$ and computing the respectivve $\gamma_c$. The resulting mean and variance are respectively $\bar{\gamma_c}=5.626, \text{Var}(\gamma_c)=0.033$, indicating a tight concentration around the mean. 


\subsection{Derivation of Eq.\ (\ref{eq:sgns_update})}\label{app:sgns_update}
Here, we show that vanilla Skip-gram with the cross-entropy loss where the target distribution is represented as a one-hot vector induces an implicit pulling action on co-occuring words and pushes away other words.

For a given context word $c$, the cross-entropy loss is:
\begin{align*}
H(p(\cdot|c), \hat{p}(\cdot|c)) = -\sum_{w\in W} \hat{p}(w|c)\log p(w|c),
\end{align*}
where $p(w|c) = \frac{e^{v_c ^\intercal u_w}}{\sum_{w' \in W} e^{v_c ^\intercal u_{w'}} }$ denotes the predicted distribution by Skip-gram. $\hat{p}(\cdot|c)$ denotes the target distribution where:
\begin{align*}
  \forall w \in W: \hat{p}(w|c) =
    \begin{cases}
      1 & \text{if $w$ is the target word}\\
      0 & \text{otherwise}
    \end{cases}       
\end{align*}
By construction of $\hat{p}(w|c)$, each term in the sum of the cross-entropy loss reduces to:
\begin{align*}
  &\hat{p}(w|c)\log p(w|c) =\\
    &\begin{cases}
      -\log \frac{e^{v_c ^\intercal u_w}}{\sum_{w' \in W} e^{v_c ^\intercal u_{w'}} } & \text{if $w$ is the target word}\\
      0 & \text{otherwise}
    \end{cases}       
\end{align*}

Thus, for a fixed context word $c$ and target word $w$, the loss of Skip-gram reduces to:
$$
\mathcal{L}_{\textup{SGNS}}(c, w) = - \log \frac{e^{v_c ^\intercal u_w}}{\sum_{w' \in W} e^{v_c ^\intercal u_{w'}}}.
$$ 

Now, consider two words $c, w$ that co-occur. Without loss of generality, if we assume $w$ appears prior to $c$ in the training corpus, Skip-gram first updates the context and target vectors of $w$ and $c$ respectively. Taking the gradient of $\mathcal{L}_{\textup{SGNS}}$ with respect to $v_a$ and $u_b$ for two co-occurring words $a, b\in W$,
\begin{align}
\frac{\partial \mathcal{L}_{\textup{SGNS}}}{\partial v_a} &= \sum_{w \in W} \left(\frac{e^{v_a ^\intercal u_b}}{\sum_{w' \in W} e^{v_a ^\intercal u_{w'}}} u_w\right) - u_b\label{eq:sgns_1},\\
\frac{\partial \mathcal{L}_{\textup{SGNS}}}{\partial u_b} &= \left( \frac{e^{v_a ^\intercal u_b}}{\sum_{w' \in W} e^{v_a ^\intercal u_{w'}}} - 1\right)v_a\label{eq:sgns_2}.
\end{align}

Observe that the gradients induce a pulling action between the vectors $v_a$ and $u_b$. 

Define the set of words that lie between $c$ and $w$ in the training corpus as $C$. Notice that $c$ and $w$ will co-occur with $\Delta-1$ words. Hence, for each word $c' \in C = \{c_1, ..., c_{w-1}\}$, the gradient update in Eq.\ (\ref{eq:sgns_1}) and (\ref{eq:sgns_2}) is applied to all word pairs $(w, c_1), ..., (w, c_{w-1})$ and $(c, c_1), ..., (c, c_{w-1})$. 

Consider the pulling action induced by the word pairs $(w, c_i)$ and $(b, c_i)$ for some $i \in [\Delta-1]$. As we first update the context and target vectors for $w$ and $c_i$, notice that 
\begin{align}
v_w^{\textup{new}} &= v_w + u_{c'} - \sum_{x \in W} \left(\frac{e^{v_w ^\intercal u_{c'}}}{\sum_{{w'} \in W} e^{v_w ^\intercal u_{w'}}} u_x\right),\nonumber\\
u_{c'}^{\textup{new}} &= u_{c'} + \left(1-\frac{e^{v_w ^\intercal u_{c'}}}{\sum_{{w'} \in W} e^{v_w ^\intercal u_{w'}}} \right)v_w\label{eq:sgns_3}.
\end{align}
Similarly, if we now update the context and target vectors for $c$ and $c_i$, 
\begin{align*}
v_c^{\textup{new}} &= v_c + u_{c'}^{\textup{new}} - \sum_{x \in W} \left(\frac{e^{v_c ^\intercal u_{c'}}}{\sum_{{w'} \in W} e^{v_c ^\intercal u_{w'}}} u_x\right).
\end{align*}
Plugging the expression for $u_c^{\textup{new}}$ in Eq.\ (\ref{eq:sgns_3}), we get
\begin{align*}
v_c^{\textup{new}} = & v_c + \left(1-\frac{e^{v_w ^\intercal u_{c'}}}{\sum_{{w'} \in W} e^{v_w ^\intercal u_{w'}}} \right)v_w + u_{c'} \\
&-\sum_{x \in W} \left(\frac{e^{v_w ^\intercal u_c}}{\sum_{{w'} \in W} e^{v_w ^\intercal u_{w'}}} u_x\right).
\end{align*}
The expression above indicates that $v_c$ is pulled towards $v_w$ implicitly and shifted closer to $u_{c'}$ explicitly in the update process while pushing away the weighted average of all word vectors. This update resembles the push-pull action in CWM. 

\subsection{Derivation of Eq.\ (\ref{eq:glove_update})}\label{app:glove_update}
For a fixed word pair $i,j$, GloVe's local objective is:
\begin{align*}
\mathcal{L}_{\textup{GloVe}}(i, j) = f(X_{ij}) (v_i^\intercal u_j + b_i + \tilde{b}_j - \log X_{ij}),
\end{align*}
where $X_{ij}$ is the co-occurrence count of words $i$ and $j$, $f(X_{ij})$ is a weighting term, $b_i, \tilde{b}_j$ are bias terms, and $v_i$, $u_j$ denote the word vector and context word vectors respectively (cf.\ \citealp{pennington-etal-2014-glove}). Typically, $f(X_{ij})$ is set to $\min\{(X_i/X_{\textup{max}})^{\alpha}, 1\}$ where $X_i$ denotes the occurrence count of word $i$ and $X_{\textup{max}} = 100$. For the sake of demonstrating the pushing action in the gradient update, we consider a weighting function $f(X_{ij}) = \min\{(X_i/X_{\textup{max}})^{\alpha} + \epsilon, 1\}$ for a arbitrarily small $\epsilon > 0$. 

The derivative of the local objective with respect to $v_i$ and $u_j$ are:
\begin{align}
\frac{\partial \mathcal{L}_{\textup{GloVe}}}{\partial v_i} &= 2f(X_{ij}) (v_i^\intercal u_j + b_i + \tilde{b}_j - \log X_{ij}) u_j,\nonumber\\
\frac{\partial \mathcal{L}_{\textup{GloVe}}}{\partial u_j} &= 2f(X_{ij}) (v_i^\intercal u_j + b_i + \tilde{b}_j - \log X_{ij}) v_i. \label{eq:glove_gradient}
\end{align}
Consider two co-occurring words $c, w$ and a word $w'$ that co-occurs with neither. Then, there exists a word $c'$ that co-occurrs with $c$ and $w$ but does not co-occur with $w'$. Define $X_{c'w'} = 0, X_{cc'} = \omega_c, X_{wc'} = \omega_w$ where $\omega_c, \omega_w\in\mathbb{N}$.

With Eq. \eqref{eq:glove_gradient}, the updated vectors for $c, w, w'$ can be written as:
\begin{align*}
&\resizebox{\hsize}{!}{$
v_c^{\textup{new}} = v_c^{\textup{old}} + 2f(\omega_c) (v_c^\intercal u_{c'} + b_c + \tilde{b}_{c'} - \log \omega_c) u_{c'}$}, \\
&\resizebox{1\hsize}{!}{$
v_w^{\textup{new}} = v_w^{\textup{old}} + 2f(\omega_w) (v_w^\intercal u_{c'} + b_w + \tilde{b}_{c'} - \log \omega_w) u_{c'}$}, \\
&\resizebox{1\hsize}{!}{$
v_{w'}^{\textup{new}} = v_{w'}^{\textup{old}} + 2f(\epsilon) (v_{w'}^\intercal u_{c'} + b_{w'} + \tilde{b}_{c'} - \log \epsilon) u_{c'} $}.
\end{align*}
As $\forall i, j: f_{X_{ij}} > 0$, notice that 
\begin{align*}
(v_c^\intercal u_{c'} + b_c + \tilde{b}_{c'} - \log \omega_c) &< 0,\\
(v_w^\intercal u_{c'} + b_w + \tilde{b}_{c'} - \log \omega_w) &< 0,\\
(v_{w'}^\intercal u_{c'} + b_{w'} + \tilde{b}_{c'} - \log \epsilon) &> 0,
\end{align*}
for sufficiently large $\omega_c$ and $\omega_w$ and for sufficiently small $\epsilon$. Setting $2\cdot |f(X_{ij}) (v_i^\intercal u_{j} + b_i + \tilde{b}_{j} - \log X_{ij})| = g(i, j)$, we see that
\begin{align}
v_c^{\textup{new}} &= v_c^{\textup{old}} + g(c, c')v_{c'},\nonumber\\
v_w^{\textup{new}} &= v_w^{\textup{old}} + g(w, c')v_{c'}\nonumber,\\
v_{w'}^{\textup{new}} &= v_{w'}^{\textup{old}} - g(w', c')v_{c'}. \nonumber
\end{align}
This indicates that $v_c$ and $v_w$ will be pulled towards the context word vectors of words that $c$ and $w$ both co-occur with, while words that do not co-occur with $c$ and $w$ will be pushed away from $v_c$ and $v_w$. 

\section{Supplementary Experiments}

\subsection{Metric Details}\label{app:metrics}
Given a set of word pairs in an analogy $A=\{(a_1, b_1), (a_2, b_2), \dots\}$, PCS measures relative directional alignment by computing the separability of cosine similarities between true vector offsets $v_{a_i} - v_{b_i}$ and false offsets $v_{a_i} - v_{b_j}, i \neq j$. Concretely, denoting the set of true and false offsets as $P$ and $N$ respectively, PCS computes the expectation of the ROC-AUC score between $P$ and a subset of the false vector offsets $N' \subset N$ where $|P| = |N'|$:
$$
\text{PCS}(A) = \mathbb{E}_{N' \sim U(N)} \left[\text{AUC}(P, N')\right],
$$
where $U(N)$ denotes the uniform distribution over all false vector offsets. Typically, the expectation is approximated by sampling $s = 50$ subsets.

In contrast, MSM represents the absolute alignment within analogies by computing the cosine similarities between all true vector offsets and the mean of the true offsets.  
$$
\text{MSM}(A) = \frac{1}{|P|} \sum_{v_p \in P} \cos\left(v_p, \frac{1}{|P|} \sum_{v_p \in P} v_p\right)
$$
A high value of MSM indicates alignment between true vector offsets. However, note that MSM is susceptible to scoring undesirable vector structures with high values (e.g.\ when all vectors are collapsed onto one point in the embedding space, MSM $ = 1$).

\begin{table}[t]
\centering
\begin{tabular}{l|c|ccc}
\hline
\textbf{Model} & \textbf{\parallelogramm} & \textbf{WordSim} & \textbf{MEN} & \textbf{SimLex}\\
\hline
\rowcolor{Gray}
CWM & 0.27 & 0.66 & 0.73 & 0.34\\
SGNS & 0.29 & 0.72 & 0.74 & 0.36 \\
GloVe & 0.29 & 0.61 & 0.75 & 0.37\\
\hline
\end{tabular}
\small{\caption{\label{table:performances_sup} Performances for embedding models on parallelogram analogy recovery and word similarity tasks. $\parallelogramm$ refers to parallelogram recovery task. For word similarity, reported values are Spearman's rank correlation between word similarity rankings of human annotators and cosine similarites computed from word vectors.}}
\end{table}

\subsection{Word Similarity}\label{app:word-sim}
While the analogy task is our primary focus, we evaluate CWM on other commonly used benchmarking tasks for completeness. To this end, we benchmark our model on WordSim353 \cite{Finkelstein2002PlacingSI}, the MEN Test Collection \cite{Bruni2014MultimodalDS}, and SimLex999 \cite{hill-etal-2015-simlex}. 

On all tasks, CWM performs comparably with existing models (Table \ref{table:performances_sup}). We highlight that minor performance differences on word similarity tasks are negligible, as such benchmarks are built using human annotations and are subject to noise. Nevertheless, we believe further refinement of the CWM model is required to boost performance on various downstream tasks. 

\subsection{Analogies as Parallelograms}\label{app:analogy_parallelograms}
We also benchmark all models on the traditional parallelogram analogy recovery task using the BATS dataset. 

Concretely, given an analogy pair $a:b=c:d$, we utilize the most common metric where we compute the $x$ that satisfies:
$$
x = \min_{x \in W\setminus\{a, b, c\}} \|{v_b} - {v_a} + {v_c} - {v_x}\|,
$$
and compare whether $x = d$. 

Results from Table \ref{table:performances_sup} indicate that CWM recovers analogies as parallelograms comparably to existing models. 

\end{document}